\pdfoutput=1

\documentclass[11pt]{article}

\usepackage[]{acl}

\usepackage{times}
\usepackage{latexsym}
\usepackage{graphicx}
\usepackage{amsfonts}
\usepackage{amsmath}
\usepackage{multirow}

\usepackage[T1]{fontenc}

\usepackage[utf8]{inputenc}

\usepackage{microtype}

\newcommand \footnoteONLYtext[1]{
    \let \mybackup \thefootnote
    \let \thefootnote \relax
    \footnotetext{#1}
    \let \thefootnote \mybackup
    \let \mybackup \imareallyundefinedcommand
}

\title{THiFLY Research at SemEval-2023 Task 7: A Multi-granularity System for CTR-based Textual Entailment and Evidence Retrieval}

\author{Yuxuan Zhou$^{1*}$, Ziyu Jin$^{2,3*}$, Meiwei Li$^{2,3}$, Miao Li$^{1}$, Xien Liu$^{1}$, Xinxin You$^{2,3}$, Ji Wu$^{1}$ \\
	$^1$Department of Electronic Engineering, Tsinghua University, Beijing 100084, China  \\
	$^2$THiFLY Research, Tsinghua University, Beijing 100084, China  \\
	$^3$State Key Laboratory of Cognitive Intelligence, Hefei, Anhui 230088, China  \\
 }
\begin{document}
\maketitle
\footnoteONLYtext{*: equal contribution}
\begin{abstract}
The NLI4CT task aims to entail hypotheses based on Clinical Trial Reports (CTRs) and retrieve the corresponding evidence supporting the justification. This task poses a significant challenge, as verifying hypotheses in the NLI4CT task requires the integration of multiple pieces of evidence from one or two CTR(s) and the application of diverse levels of reasoning, including textual and numerical. To address these problems, we present a multi-granularity system for CTR-based textual entailment and evidence retrieval in this paper. Specifically, we construct a \textbf{Mul}ti-\textbf{g}ranularity Inference \textbf{Net}work (\textbf{MGNet}) that exploits sentence-level and token-level encoding to handle both textual entailment and evidence retrieval tasks. Moreover, we enhance the numerical inference capability of the system by leveraging a T5-based model, SciFive, which is pre-trained on the medical corpus. Model ensembling and a joint inference method are further utilized in the system to increase the stability and consistency of inference. The system achieves f1-scores of \textbf{0.856} and \textbf{0.853} on textual entailment and evidence retrieval tasks, resulting in the best performance on both subtasks. The experimental results corroborate the effectiveness of our proposed method. Our code is publicly available at \url{https://github.com/THUMLP/NLI4CT}.
\end{abstract}

\section{Introduction}
In recent years, with the fast development of digital health, there has been a surge in the publication of Clinical Trial Reports (CTRs). Currently, there are over 10,000 CTRs for Breast Cancer alone\footnote{\url{https://sites.google.com/view/nli4ct/home}}. This proliferation of CTRs has enabled the construction of a Natural Language Inference (NLI) system that can aid medical interpretation and evidence retrieval for personalized evidence-based care. The NLI4CT task \citep{jullien-2023-nli4ct} focuses on constructing an explainable multi-evidence NLI system based on CTRs for Breast Cancer. It consists of two sub-tasks: textual entailment (Task A) and evidence retrieval (Task B). Task A requires the system to determine if the CTRs entail or contradict the given hypotheses. Task B requires the system to find evidence in the CTRs necessary to justify the prediction in Task A. 
The primary difficulty inherent in NLI4CT is that the process of hypothesis verification frequently involves the integration of several pieces of evidence from the premise. In certain cases, justifying a hypothesis requires a comparison of two separate premise CTRs. Moreover, unlike other clinical NLI benchmarks (e.g., MedNLI), NLI4CT presents a unique feature containing four distinct types of premises - Intervention, Eligibility, Results, and Adverse Events - that align with the four sections typically found in each CTR. Validating hypotheses based on each premise type requires different levels of inference skills (textual, numerical, etc.). 
\begin{figure}[t]
	\centering
	\includegraphics[scale=0.57]{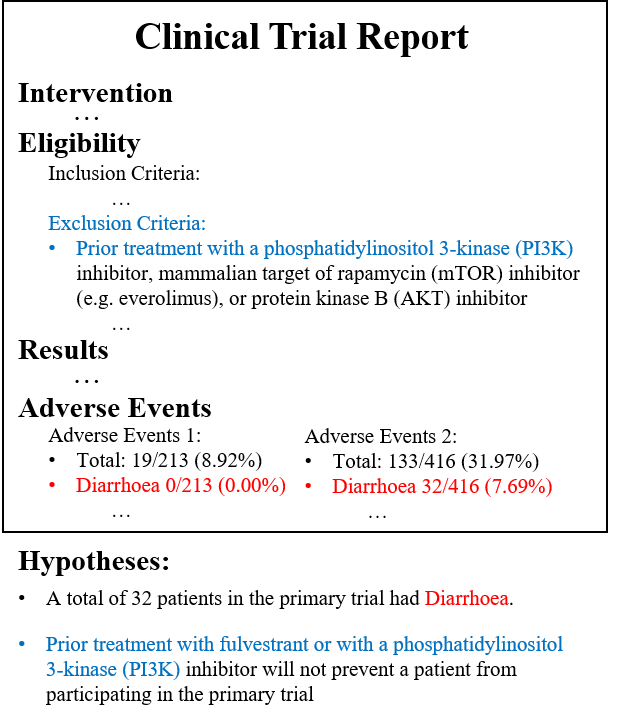}
	\caption{An Example of NLI4CT.}
	\label{fig:verification example}
\end{figure}
\begin{figure*}[t]
	\centering
	\includegraphics[width=\textwidth]{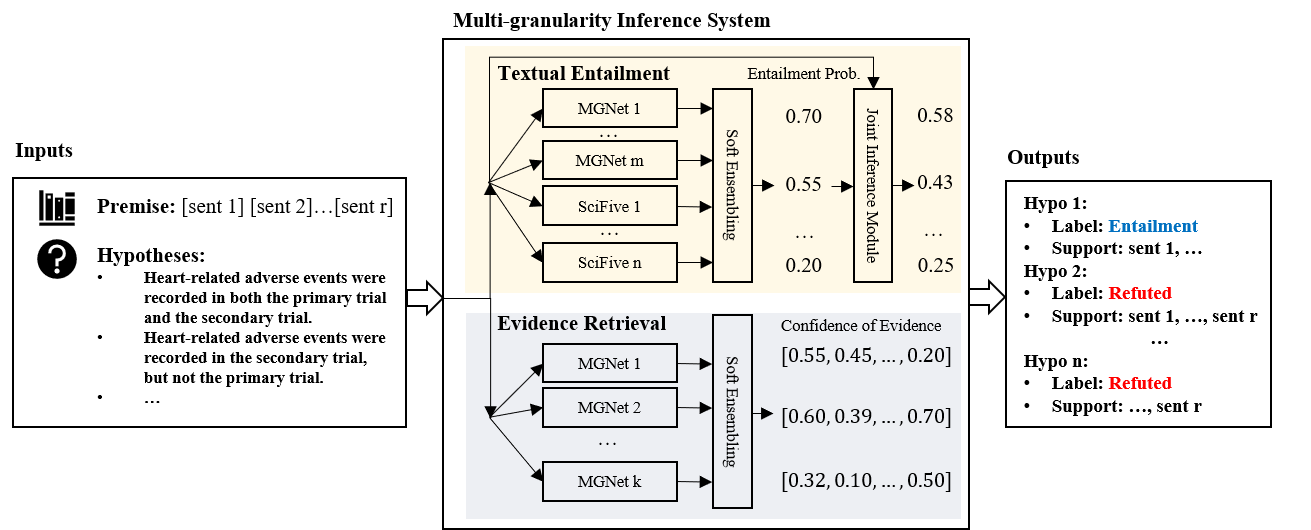}
	\caption{An overview of the proposed Multi-granularity System for NLI4CT task. MGNet refers to the proposed multi-granularity inference network.}
	\label{fig:system}
\end{figure*}
Fig.\ref{fig:verification example} presents an example of NLI4CT. The first hypothesis involves summating the numbers of patients who experienced Diarrhoea in two distinct cohorts to conduct the verification. Alternatively, verifying the second hypothesis requires evaluating treatment modality against the criteria outlined in the Exclusion Criteria.

This paper proposes a multi-granularity inference system for the NLI4CT task to address the aforementioned challenges. An overview of the proposed system is presented in Fig.\ref{fig:system}. The system comprises a multi-granularity inference network that performs joint semantics encoding of the hypothesis and premise, followed by multi-granularity inference through sentence-level and token-level encoding of the encoded sequence to accomplish Task A and Task B. To handle the various levels of inference involved, we utilize the SciFive model \citep{phan2021scifive} to enhance the system's performance on hypotheses requiring numerical inference skills. Our system is presented as an ensemble of models to increase the stability and consistency of inference. A joint inference module is further proposed to refine the inference outcomes of hypotheses by detecting semantic consistency between hypothesis pairs sharing the same premise. Our system achieves micro f1-scores of 0.856 and 0.853 on Task A and Task B, respectively. The main contributions of this work are as follows:
\begin{itemize}
\item We present a multi-granularity inference network that performs both textual entailment and evidence retrieval tasks concurrently. We further incorporate the SciFive model to strengthen the system's numerical inference capability.
\item We introduce a joint inference module that improves the inference consistency by exploiting the polarity of hypotheses pairs that share the same premise.
\item Our proposed framework achieves state-of-the-art f1-scores of 0.856 and 0.853 on Task A and B, respectively.
\end{itemize}

\section{Related Work}
\paragraph{BiolinkBERT}
BiolinkBERT \citep{yasunaga-etal-2022-linkbert} is a pre-trained model that utilizes hyperlinks within documents. The model incorporates linked documents into the input of pre-training and is pre-trained using masked language modeling and document relation prediction tasks. BiolinkBERT has demonstrated state-of-the-art performance on various medical NLP benchmarks, such as BLURB \citep{gu2021domain} and BioASQ \citep{nentidis2020results}.
\paragraph{SciFive}
SciFive \citep{phan2021scifive} is a T5-based model pre-trained on large biomedical corpora, namely PubMed Abstract and PubMed Central. This model has shown the potential of text-generating models in medical NLP tasks and has achieved state-of-the-art performance on MedNLI \citep{romanov2018lessons} and DDI \citep{herrero2013ddi}.
\section{Task Formulation}
The Multi-evidence Natural Language Inference for Clinical Trial Data (NLI4CT) task comprises two sub-tasks: textual entailment and evidence retrieval. Both sub-tasks share a common input comprising a premise $P$ and a hypothesis $S$. The hypothesis is a single sentence, while the premise comprises several sentences in the CTR(s), i.e., $P=[P^1, P^2, \cdots, P^m]$.

The textual entailment sub-task can be viewed as a binary classification problem, with the entailment module $f_{\mathrm{Ent}}$ generating a predicted result $\hat{\textbf{y}}\in\{0,1\}^n$ for all input hypotheses: $f_{\mathrm{Ent}}(S, P)=\hat{y}$. In this context, the predicted value $\hat{y}=1$ indicates that the premise entails the hypothesis, while $\hat{y}=0$ denotes that the hypothesis contradicts the premise. On the other hand, the evidence retrieval sub-task can be seen as a binary classification problem targeting each sentence in the premise. The evidence retrieval module $f_{\mathrm{Ret}}$ produces a predicted result $\hat{\textbf{r}}\in\{0,1\}^m$: $f_{\mathrm{Ret}}(S, P)=\hat{\textbf{r}}$, where $\hat{r}_j=1$ indicates that the $j^{th}$ sentence in the premise supports the justification of $S$, $\hat{r}_j=0$ means the sentence is not related to the justification of $S$.
\section{Models}
In this section, we introduce models applied in our inference system: \textbf{M}ulti-\textbf{g}ranularity Inference \textbf{Net}work (\textbf{MGNet}), SciFive \citep{phan2021scifive}, and the joint inference network. 
\begin{figure}[t]
	\centering
	\includegraphics[scale=1]{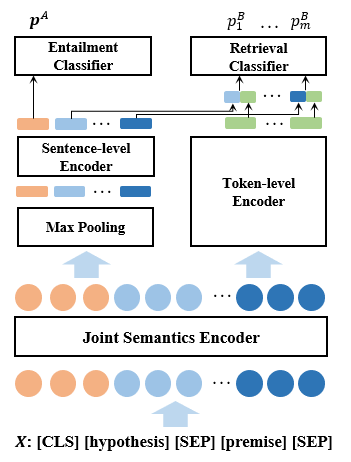}
	\caption{The structure of the proposed Multi-granularity Inference Network (MGNet).}
	\label{fig:mgnet}
\end{figure}
\subsection{Multi-granularity Inference Network}
The multi-granularity inference network aims to handle both the textual entailment and evidence retrieval subtasks using integrated token-level and sentence-level representations. The network structure is illustrated in Figure \ref{fig:mgnet}. This section is organized as follows: Section \ref{sec: joint} presents the joint semantics encoder, which encodes the contextual semantics between the hypothesis and the premise. Section \ref{sec: sent} describes the sentence-level encoder, which learns the sentence-level contextual semantics of the hypothesis and the premise. Section \ref{sec: tok} describes the token-level encoder, which extracts the token-level joint representation between the hypothesis and each sentence in the premise. Finally, Section \ref{sec: cls} introduces the classifier module, which accomplishes Task A and Task B inference based on the learned multi-granularity semantics.
\subsubsection{Joint Semantics Encoder}\label{sec: joint}
The network initially employs a transformer-based \citep{vaswani2017attention} language model to learn the joint contextual representation of the hypothesis and premise, aiming to incorporate medical domain knowledge into the network. Specifically, we concatenate each hypothesis $S$ with the premise $P$ to form an input token sequence $\textbf{X}=[[\mathrm{CLS}],\textbf{S},[\mathrm{SEP}],\textbf{P},[\mathrm{SEP}]]$, where $\textbf{S}$ and $\textbf{P}$ refer to the tokenization of $S$ and $P$, respectively. Further details regarding data preprocessing can be found at Sec.\ref{sec:input} in the appendix. The token sequence is subsequently embedded and encoded by the transformer-based language model as follows: 
\begin{equation}
    \textbf{H}^{0}=f_{Emb}(\textbf{X})
\end{equation}
\begin{equation}
    \textbf{H}^{l}=f_{Enc}^{l}(\textbf{H}^{l-1})
\end{equation}
where $\textbf{H}^0$ denotes the output of the embedding layer, $\textbf{H}^{l} \in \mathbb{R}^{N\times d} $ $(l\in\{1,2,\cdots,L\})$ refers to the output representation of the $l^{th}$ transformer encoder layer, $N$ is the sequence length and $d$ the dimension of the representation vector. Here, $f_{Emb}$ and $f_{Enc}^{l}$ refer to the embedding layer and the $l^{th}$ encoding layer, respectively. Finally, the output representation of the final layer, $\textbf{H}^{L}$, is further processed in the subsequent encoders.
\subsubsection{Sentence-level Encoder}\label{sec: sent}
The sentence-level encoder conducts contextual encoding of the hypothesis and the premise based on the learned token-level representation. Specifically, the encoder first conducts max-pooling\footnote{We have tried other pooling methods (mean-pooling, attention pooling, etc.). However, they fail to achieve higher performance than max-pooling.} to extract sentence-level representation from the token-level representation of each sentence, as follows: 
\begin{equation}
    \hat{\textbf{H}}^{s}_i=\operatorname{Max-pooling}(\{\textbf{H}^{L}_j: j\in \textbf{I}_{i}\})
\end{equation}
where $\hat{\textbf{H}}^{s}\in \mathbb{R}^{(m+1)\times d}$ refers to the pooled sentence-level representation and $\textbf{I}_{i}$ denotes the subscript set of $\textbf{X}$ that corresponds to the tokens in the $i^{th}$ sentence. For example, $\hat{\textbf{H}}^{s}_1$ corresponds to the hypothesis, while $\hat{\textbf{H}}^{s}_{i+1}$ corresponds to the ${i}^{th}$ sentence in the premise. 

After that, the resulting $\hat{\textbf{H}}^{s}$ output is fed into an encoder designed to capture contextual semantics. The motivation for this encoder stems from the need to justify some hypotheses using multiple pieces of evidence. To achieve this, we implement the encoder by two alternative networks: 

(1) BiLSTM \citep{hochreiter1997long}: the network utilizes two LSTMs to compute the representation of the given sequence from both directions: 
\begin{equation}
    \overset{\rightarrow}{\textbf{H}}^{s} = \operatorname{LSTM}^{s}_{for}(\hat{\textbf{H}}^{s})
\end{equation}
\begin{equation}
    \overset{\leftarrow}{\textbf{H}}^{s} = \operatorname{LSTM}^{s}_{rev}(\hat{\textbf{H}}^{s})
\end{equation}
The final representation of each sentence $\textbf{H}^{s}\in\mathbb{R}^{(m+1)\times d}$ is obtained by concatenating the forward and backward representations and passing them to a linear layer:
\begin{equation}
    \textbf{H}^{s}_i = \textbf{W}_1[\overset{\rightarrow}{\textbf{H}}^{s}_i;\overset{\leftarrow}{\textbf{H}}^{s}_i]+\textbf{b}_1
\end{equation}
where $\textbf{W}_1\in\mathbb{R}^{d\times 2d}, \textbf{b}_1\in\mathbb{R}^{d}$ are trainable parameters. Finally, the global representation of the hypothesis-premise pair $\Tilde{\textbf{h}}^{s}\in\mathbb{R}^{d}$ is computed by concatenating the last hidden state of the forward and backward encoding ($\textbf{W}_2\in\mathbb{R}^{d\times 2d}, \textbf{b}_2\in\mathbb{R}^{d}$ are also trainable parameters):
\begin{equation}
    \Tilde{\textbf{h}}^{s} = \textbf{W}_2[\overset{\rightarrow}{\textbf{H}}^{s}_{m+1};\overset{\leftarrow}{\textbf{H}}^{s}_1]+\textbf{b}_2
\end{equation}

(2) Transformer \citep{vaswani2017attention}: a $L^s$-layer transformer encoder integrates reasoning-related evidence with the sentence-level attention mechanism. In this case, we take the last hidden state as the final representation of each sentence and the hypothesis's representation as the global representation:
\begin{equation}
    \textbf{H}^{s} = f^{s}_{Enc}(\hat{\textbf{H}}^{s})
\end{equation}
\begin{equation}
    \Tilde{\textbf{h}}^{s} = \textbf{H}^{s}_i
\end{equation}
\subsubsection{Token-level Encoder}\label{sec: tok}
The purpose of the token-level encoder is to capture the reasoning-related semantics of each sentence in the premise. Compared to the sentence-level encoder,  it provides a fine-grained representation of individual sentences in premise, which has aided in evidence retrieval. To encode the $i^{th}$ sentence in the premise, we concatenate it with the hypothesis and input it into the encoder. Two alternative networks can be used to implement the encoder:

(1) BiLSTM: A BiLSTM network is leveraged to encode the forward and backward representation of the given sequence:
\begin{equation}
    \overset{\rightarrow}{\textbf{H}}^{t}_i = \operatorname{LSTM}^{t}_{for}([\textbf{H}^L_{\textbf{I}_1},\textbf{H}^L_{\textbf{I}_{i+1}}])
\end{equation}
\begin{equation}
    \overset{\leftarrow}{\textbf{H}}^{t}_i = \operatorname{LSTM}^{t}_{rev}([\textbf{H}^L_{\textbf{I}_1},\textbf{H}^L_{\textbf{I}_{i+1}}])
\end{equation}
where $\textbf{H}^L_{\textbf{I}_j}\in \mathbb{R}^{N_j\times d}$ refers to the representation of tokens from the $j^{th}$ sentence in $\textbf{X}$, $N_j$ denotes the length of the $j^{th}$ sentence, $[\cdot,\cdot]$ denotes the concatenation of matrices on the first dimension, $\overset{\rightarrow}{\textbf{H}}^{t}_i, \overset{\leftarrow}{\textbf{H}}^{t}_i\in\mathbb{R}^{(N_1+N_{i+1})\times d}$ are the output representation of the LSTMs. Finally, the representation of the $i^{th}$ sentence in the premise is also obtained by concatenating the last hidden state of LSTMs:
\begin{equation}
    \textbf{H}^t_i = \textbf{W}_3[\overset{\rightarrow}{\textbf{H}}^{t}_{N_1+N_{i+1}};\overset{\leftarrow}{\textbf{H}}^{t}_1]+\textbf{b}_3
\end{equation}
where $\textbf{W}_3\in\mathbb{R}^{d\times 2d}$ and $\textbf{b}_3\in\mathbb{R}^{d}$ are trainable parameters as well.

(2) max-pooling: A simpler implementation of the encoder is a max-pooling layer that merges the representation of the sequence:
\begin{equation}
    \textbf{H}^t_i = \operatorname{Max-pooling}(([\textbf{H}^L_{\textbf{I}_1},\textbf{H}^L_{\textbf{I}_{i+1}}]))
\end{equation}
\subsubsection{Classifiers}\label{sec: cls}
We implement the classifiers with simple structures for both Task A and B. For Task A (textual entailment), we use a double-layer MLP that takes the sentence-level global representation $\Tilde{\textbf{h}}^s$ as the input:
\begin{equation}
    \textbf{p}^A = \operatorname{Softmax}(\textbf{W}^A_2\sigma(\textbf{W}^A_1\Tilde{\textbf{h}}^s+\textbf{b}^A_1)+\textbf{b}^A_2)
\end{equation}
where $ \textbf{W}^A_1\in\mathbb{R}^{d\times d},\textbf{b}^A_1\in\mathbb{R}^{d},\textbf{W}^A_2\in\mathbb{R}^{2\times d},\textbf{b}^A_2\in\mathbb{R}^{2}$ are MLP's parameters, $\sigma$ denotes the GELU activation function \citep{hendrycks2016gaussian}. The output $\textbf{p}^A$ suggests that the given hypothesis has a $p^A_1$ probability of being false and a $p^A_2$ probability of being true based on the premise. For Task B (evidence retrieval), we project the concatenation of sentence-level and token-level representations into a scalar value to determine whether the $i^{th}$ sentence in the premise supports the hypothesis:
\begin{equation}
    p^B_i = \operatorname{Sigmoid}(\textbf{W}^B[\textbf{H}^s_i;\textbf{H}^t_i]+b^B)
\end{equation}
where $\textbf{W}^B\in\mathbb{R}^{1\times 2d}, b^B\in\mathbb{R}$ are trainable parameters. $p^B_i$ measures the probability that the $i^{th}$ sentence in the premise supports the justification of the hypothesis. 
\subsection{SciFive}
The proposed MGNet performs well on Task A and Task B through a unified multi-granularity encoding process. Nevertheless, we found in experiments that MGNet struggles with hypotheses that require numerical inference skills. Motivated, we implement SciFive to enhance the system's numerical inference ability. Following the original paper, we preprocess the input sequence $\textbf{X}$ in the format of ``nli hypothesis: [hypothesis] premise: [premise]". SciFive then computes the probabilities of generating ``entailment" ($P(``ent"|\textbf{X})$) and ``contradiction" $P(``con"|\textbf{X})$ given the input sequence. The prediction for Task A is then calculated using the following equations:
\begin{equation}
    p^A_1 = \frac{P(``con"|\textbf{X})}{P(``con"|\textbf{X})+P(``ent"|\textbf{X})}
\end{equation}
\begin{equation}
    p^A_2 = \frac{P(``ent"|\textbf{X})}{P(``con"|\textbf{X})+P(``ent"|\textbf{X})}
\end{equation}
\subsection{Joint Inference Network}\label{sec: joint inference}
The inference models we implemented have demonstrated satisfactory performance in justifying single hypotheses. However, these models tend to produce identical predictions for hypotheses that share the same premise but contradict each other, leading to inaccurate results. For example, consider the hypotheses ``Heart-related adverse events were recorded in \textbf{both} the primary trial \textbf{and} the secondary trial" and ``Heart-related adverse events were recorded in the secondary trial, \textbf{but not} the primary trial," which are semantically exclusive. Assigning the same label to both hypotheses is an unreasonable outcome. It is worth noting that such a problem will likely occur in real-world evidence-based care systems, where users may interact with the system multiple times and generate multiple hypotheses based on the same premise.

Motivated by this, we aim to enhance inference consistency and performance by leveraging the mutual information among hypotheses that share the same premise. To achieve this, we consider a set of hypotheses $S^1, S^2,\cdots, S^n$ that share the same premise and design a joint inference network that determines whether a given pair of hypotheses has the same label for Task A under the given premise. We implement the network with a $L$-layer transformer encoder. The network initially takes the sequence $\Bar{\textbf{X}}=[[\mathrm{CLS}],\textbf{S}^i,[\mathrm{SEP}], \textbf{S}^j,[\mathrm{SEP}],\textbf{P},[\mathrm{SEP}]]$ as input:
\begin{equation}
    \Bar{\textbf{H}}^{L} = f^{J}_{Enc}(\Bar{\textbf{X}})
\end{equation}
where $f^{J}_{Enc}$ refers to the $L$-layer transformer encoder. A double-layer MLP is applied to make predictions based on the last hidden state of $[\mathrm{CLS}]$ that is considered as the representation vector of the entire sequence:
\begin{equation}
    \textbf{c}^{i,j} = \operatorname{Softmax}(\textbf{W}^J_2\sigma(\textbf{W}^J_1\Bar{\textbf{H}}^{L}+\textbf{b}^J_1)+\textbf{b}^J_2)
\end{equation}
where $ \textbf{W}^J_1\in\mathbb{R}^{d\times d},\textbf{b}^J_1\in\mathbb{R}^{d},\textbf{W}^J_2\in\mathbb{R}^{2\times d},\textbf{b}^J_2\in\mathbb{R}^{2}$ are trainable parameters and $\sigma$ represents the $\operatorname{Tanh}$ activation function. $\textbf{c}^{i,j}_1$ measures the probability that $S^i$ and $S^j$ share the same label, and $\textbf{c}^{i,j}_2$ measures the probability of having different labels. The joint inference module rectifies the predictions of Task A with the prediction results of this network, and we will introduce the details in the next section.
\section{System Details}
\paragraph{Model Setup}
We initialized the joint semantics encoder in MGNet and the joint inference network with BiolinkBERT-Large-MNLI-SNLI, a language model proposed by \citet{yasunaga-etal-2022-linkbert} that is pre-trained on the PubMed dataset and further finetuned on the Multi-Genre Natural Language Inference (MNLI) \citep{N18-1101} and Stanford Natural Language Inference (SNLI) datasets \citep{bowman-etal-2015-large}. Following the setting of BiolinkBERT-Large, we set the number of layers $L$ and the dimension of hidden states $d$ to 24 and 1024, respectively. We apply Adam \citep{kingma2014adam} and Adafactor optimizers \citep{shazeer2018adafactor} for finetuning MGNet and SciFive, respectively.

\begin{table*}[t]\small
\centering
\setlength{\tabcolsep}{6mm}{
\begin{tabular}{llllll}
\hline
\multirow{2}{*}{Name} & \multirow{2}{*}{Network} & \multirow{2}{*}{ML} & \multicolumn{2}{c}{Encoder}            & \multirow{2}{*}{Loss} \\ \cline{4-5}
                      &                          &                             & Sentence       & Token     &                        \\ \hline\hline
\multicolumn{6}{c}{Task A}                                                                                                                      \\ \hline\hline
M-512-Bi-Bi-mul                     & MGNet                    & 512                         & BiLSTM               & BiLSTM          & $\mathcal{L}_{mul}$                       \\
M-512-Tf-Bi-cl                     & MGNet                    & 512                         & Transformer          & BiLSTM          & $\mathcal{L}_{CL}$                       \\
M-1024-Tf-Bi-mul                     & MGNet                    & 1024                        & Transformer          & BiLSTM          & $\mathcal{L}_{mul}$                       \\
SciFive                     & SciFive                  & 1024                        & -                    & -               & $\mathcal{L}_{A}^*$                       \\ \hline\hline
\multicolumn{6}{c}{Task B}                                                                                                                      \\ \hline\hline
M-512-Tf-Bi                     & MGNet                    & 512                         & Transformer & BiLSTM & $\mathcal{L}_{B}$             \\
M-512-Bi-Bi                     & MGNet                    & 512                         & BiLSTM               & BiLSTM          & $\mathcal{L}_{B}$                      \\
M-512-Bi-Max                     & MGNet                    & 512                         & BiLSTM               & Max-pooling     & $\mathcal{L}_{B}$                     \\ \hline
\end{tabular}}
\caption{The inference models applied in the proposed system. ML: max length of the input sequence, Tf: transformer model, Bi: BiLSTM model, Max: Max-pooling. $\mathcal{L}_{A}^*$ denotes the sequence cross-entropy loss proposed in \citet{raffel2020exploring}.}
	\label{table:models}
\end{table*}
To provide a comprehensive description of the inference models used in our system, we first outline the loss functions employed in model training:

(1) Cross-entropy loss for textual entailment:
\begin{equation}
    \mathcal{L}_{A} = -(1-y)\cdot\log{p^A_1}-y\cdot\log{p^A_2}
\end{equation}
Here, $y\in\{0,1\}$ represents the label of Task A.

(2) Cross-entropy loss for evidence retrieval:
\begin{equation}
    \mathcal{L}_{B} = -\frac{1}{m}\sum_{i=1}^{m}[(1-r_i)\cdot\log{(1-p^B_i)}+r_i\cdot\log{p^B_i}]
\end{equation}
where $r_i\in\{0,1\}$ is the label of the $i^{th}$ sentence in the premise. 

(3) Multitask learning loss: we use a multitask learning loss, which is a combination of the CE loss for Task A and Task B:
\begin{equation}
    \mathcal{L}_{mul} = \mathcal{L}_{A} + \lambda \mathcal{L}_{B}
\end{equation}
Here, $\lambda$ is a hyperparameter that controls the pace of learning on both tasks.

(4) Contrastive learning loss: we use the contrastive learning loss proposed by \citet{gunelsupervised}, which is a combination of the CE loss and the supervised contrastive learning (SCL) loss:
\begin{equation}
    \mathcal{L}_{CL} = \gamma \mathcal{L}_{A} + (1-\gamma) \mathcal{L}_{SCL}
\end{equation}
Here, $\mathcal{L}_{SCL}$ is the SCL loss that takes the global representation $\Tilde{\textbf{h}}_1^s,\Tilde{\textbf{h}}_2^s,\cdots,\Tilde{\textbf{h}}_N^s$ and the labels $y_1,y_2,\cdots,y_N$ in the batch as input ($N$ is the batch size). Since our dataset is relatively small, we apply the SCL loss as a regularization technique during training.

The inference models applied in the proposed system are listed in Table.\ref{table:models}. To extend the maximum input length of the original BiolinkBERT model from 512 to 1024, we append the positional embeddings of BiolinkBERT with randomly initialized parameters. In Task A, we set the number of layers $L^s$ for the transformer in the sentence-level encoder as one and the learning rates of the joint semantics encoder and other parts as 2e-5 and 1e-4, respectively. The batch size, number of epochs, and warmup ratio are set to 32, 100, and 0.3, respectively. For SciFive, we set the learning rate, batch size, number of epochs, and warmup steps as 3e-5, 32, 100, and 500, respectively. The hyperparameter values $\lambda=0.01$ and $\gamma=0.5$ are chosen, and the temperature of the supervised contrastive learning (SCL) loss is set to 0.3. In Task B, we set the learning rate, $L^s$, batch size, number of epochs, and warmup ratio as 5e-6, 2, 1, 50, and 0.05, respectively.

In the joint inference network, the learning rate, batch size, number of epochs, and warmup ratio are set to 2e-5, 32, 100, and 0.3, respectively. Positive samples, i.e., hypothesis pairs with the same label, are created using back translation, while negative samples are obtained by identifying hypothesis pairs with different labels that appear in the datasets. Further training details of this network are presented in Section \ref{sec:JIN} in the appendix, owing to space limitations.
\paragraph{Soft Ensembling}
We apply a cross-validation-based soft ensembling method to summarize the inference results from different models. Specifically, we perform 10-fold validation for each model in Table.\ref{table:models} using all data in the train and dev sets. For Task A, we save the checkpoint that achieves the best f1-score on the corresponding dev set for each fold, while for Task B, we save two of the best checkpoints for each fold. Then, we average the predictions of $4\times 10=40$ models trained by cross-validation for Task A and average the predictions of $3\times 10\times 2+3=63$ models trained by cross-validation (two checkpoints for each fold) and hold-out method (one checkpoint for each model) for soft ensembling. The ensemble predictions for the two tasks are denoted as $\Tilde{\textbf{p}}^A$ and $\Tilde{\textbf{p}}^B$, respectively.
\begin{table*}[t]
\centering
\setlength{\tabcolsep}{4mm}{
\begin{tabular}{lcccccc}
\hline
\multirow{2}{*}{system} & \multicolumn{3}{c}{Individual Inference} & \multicolumn{3}{c}{Joint Inference} \\ \cline{2-7} 
                        & recall   & precision   & f1-score        & recall & precision & f1-score       \\ \hline
original system         & 0.816    & 0.739       & 0.776           & 0.856  & 0.856     & \textbf{0.856} \\
- M-512-Bi-Bi-mul     & 0.820    & 0.732       & 0.774           & 0.840  & 0.837     & 0.838          \\
- M-512-Tf-Bi-cl      & 0.796    & 0.748       & 0.771           & 0.848  & 0.845     & 0.846          \\
- M-1024-Tf-Bi-mul    & 0.820    & 0.745       & \textbf{0.781}  & 0.848  & 0.851     & 0.850          \\
- SciFive             & 0.792    & 0.717       & 0.753           & 0.844  & 0.844     & 0.844          \\ 
\hline
BiolinkBERT-Large       & 0.712    & 0.674       & 0.693           & 0.768  & 0.768     & 0.768          \\ 
SciFive                 & 0.784    & 0.713       & 0.747           & 0.804  & 0.798     & 0.801          \\
\hline
\end{tabular}}
\caption{Performance of the proposed system on Task A with ablation experimental results compared with baseline models on the test set of NLI4CT. }
	\label{table:res TaskA}
\end{table*}
\begin{table}[t]
\centering
\setlength{\tabcolsep}{1mm}{
\begin{tabular}{lccc}
\hline
system           & recall & precision & f1-score                              \\ \hline
original system  & 0.898  & 0.811     & {\textbf{0.853}} \\
- M-512-Tf-Bi  & 0.884  & 0.818     & 0.850                                 \\
- M-512-Bi-Bi  & 0.882  & 0.801     & 0.840                                 \\
- M-512-Bi-Max & 0.895  & 0.811     & 0.852                                 \\ \hline
\end{tabular}}
\caption{Performance of the proposed system on Task B with ablation experimental results.}
	\label{table:res TaskB}
\end{table}
\paragraph{Joint Inference}
As mentioned above, we utilize a joint inference module to rectify the ensembling results $\Tilde{\textbf{p}}^A$ before performing Task A. This module corrects the prediction of hypotheses that share the same premise using the following equation:
\begin{equation}
    \hat{\textbf{p}}^{A,i} = \frac{1}{n}\sum^{n}_{j=1}(I_{ij}\Tilde{\textbf{p}}^{A,j}+(1-I_{ij})(1-\Tilde{\textbf{p}}^{A,j}))
\end{equation}
Here, $\Tilde{\textbf{p}}^{A,i}$ represents the prediction of the $i^{th}$ hypothesis after ensembling, while $c^{i,j}1$ denotes the prediction results of the joint inference network discussed in Sec.\ref{sec: joint inference}. The function $I{ij}=\mathbb{I}[c^{i,j}1>0.5]$ employs an indicative function $\mathbb{I}[\cdot]$, which is equal to 1 if the condition in brackets holds, and 0 otherwise. We set $I{ii}=1$ since two identical hypotheses must share the same labels. Notably, when $n=2$ and the two hypotheses are predicted to be contradicting each other ($c^{1,2}_1<0.5$), the hypothesis with a higher confidence level in its prediction retains its original prediction, while the hypothesis with a lower confidence level modifies its prediction. Therefore, the joint inference module improves the system's prediction consistency.
\paragraph{System Output}
We use a threshold-based decision strategy in both tasks to obtain the ultimate predictions. Specifically, for Task A, hypotheses with $\hat{p}^{A}_2>\eta_A$ are predicted as ``entailment", while the remaining hypotheses are predicted as ``contradiction." For Task B, the system determines whether the $i^{th}$ sentence in the premise supports the inference by checking if $\Tilde{p}^{B}_i>\eta_B$. In our implementation, we set the threshold values to $\eta_A=0.57$ and $\eta_B=0.53$.
\section{Evaluation}
\subsection{Data, Metrics, and Baselines}
We conduct all the evaluations on the NLI4CT dataset. The train, dev, and test sets contain 1,700, 200, and 500 hypotheses. The hypotheses can be divided by the number of CTRs involved: (1) Single, related to only one CTR; (2) Comparison, requiring the comparison between two CTRs for justification. The premise is categorized into four sections that correspond to the CTR sections: (1) Intervention; (2) Eligibility; (3) Results; (4) Adverse Events. Additional dataset details are available in Sec.\ref{sec:data details} in the appendix. Recall, precision, and f1-score are chosen as the evaluation metrics for both Task A and Task B.

We compare our system with several baseline models (BiolinkBERT-Large, SciFive) that achieve SOTA performance on medical NLI tasks. For the BiolinkBERT-Large model, we feed the concatenated hypothesis and premise sequence into the language model and add an MLP that takes the representation of $[\mathbf{CLS}]$ as input for classification. Note that the performance of baseline models is obtained by ensembling 10 models through cross-validation to make the results comparable. We did not train baseline models for Task B because it is hard to find an existing implementation that can directly be applied to Task B. 
\subsection{Overall Performance}
The results of the proposed system on Task A and B are presented in Table \ref{table:res TaskA} and \ref{table:res TaskB}, respectively. Our system performs best on both tasks with f1-scores of 0.856 and 0.853 for Task A and Task B, respectively. Compared with the baseline models, our system shows a significant improvement of approximately 3\% and 5\% on the f1-score with and without the joint inference module, respectively, demonstrating the effectiveness of our approach. Note that the proposed multi-granularity encoding method greatly improves the performance of the original BiolinkBERT-Large model on Task A (- SciFive vs. BiolinkBERT-Large) by around 6\% in f1-score (0.753 vs. 0.693, 0.844 vs. 0.768).
\subsection{Ablation Study}
We performed ablation experiments to investigate the effect of the proposed joint inference method and the contribution of each model in the soft ensembling. The results are presented in Table \ref{table:res TaskA} and \ref{table:res TaskB}. The results show that the joint inference module leads to a significant improvement of 8\% points on Task A,  suggesting that the system is more likely to make the correct prediction when considering the mutual information between hypotheses, i.e., whether the hypotheses share the same label.

\begin{table}[t]
\centering
\setlength{\tabcolsep}{4mm}{
\begin{tabular}{lcc}
\hline
Model & Dev          & Test        \\ \hline
M-512-Bi-Bi                  & \textbf{0.896}        &\textbf{0.829}             \\
- token-level        & 0.878        &            0.816 \\
- sentence-level     & 0.857        &             0.827\\ \hline
\end{tabular}}
\caption{Ablation study of the multi-granularity encoding in MGNet on Task B. All the models are trained on the train set and evaluated on the dev and test sets. The performance is reported by the f1-score.}
	\label{table:Ablation MGNet}
\end{table}

The experimental results suggest that the MGNet, which employs both sentence-level and token-level encoders implemented by BiLSTM (M-512-Bi-Bi/M-512-Bi-Bi-mul), plays a crucial role in both Task A and B, contributing to 1.8\% and 1.3\% improvements in f1-score, respectively. Furthermore, the individual inference case results reveal that SciFive provides the most significant improvement in the system's performance, with a 2.3\% increase in the f1-score.

To further investigate the impact of multi-granularity encoding in MGNet, we conduct an additional ablation experiment on Task B by removing either the sentence-level or token-level encoder (Task A only relies on the sentence-level encoder). The results in Table \ref{table:Ablation MGNet} show that both sentence-level and token-level encoders enhance the system's performance, with the sentence-level encoder contributing more to the dev set (4\% improvement) and the token-level encoder contributing more to the test set (1.3\% improvement). The results indicate the effectiveness of the proposed multi-granularity encoding.
\begin{table}[t]
\centering
\setlength{\tabcolsep}{1mm}{
\begin{tabular}{lcccc}
\hline
Model            & Int   & Elig    & Res        & AE \\ \hline
M-1024-Tf-Bi-mul & 0.749          & 0.727          & 0.735          & 0.709          \\
SciFive          & 0.748          & 0.735          & 0.712          & \textbf{0.728} \\
M-512-Bi-Bi-mul  & 0.746          & 0.739          & \textbf{0.755} & 0.708          \\
M-512-Tf-Bi-cl   & \textbf{0.783} & \textbf{0.743} & 0.724          & 0.709          \\ \hline
\end{tabular}}
\caption{Performance (f1-score) of each model on different sections of CTR in the cross-validation of Task A. Int: Intervention, Elig: Eligibility, Res: Results, AE: Adverse Events.}
	\label{table:Task A CV}
\end{table}
\subsection{Performance on Different Sections}
We further examine the behavior of each model that contributes to the ensembling by evaluating the performance of our system on different sections of CTR. The analysis results are presented in Table \ref{table:Task A CV}. It can be observed that MGNets demonstrate superior performance on hypotheses involving the Intervention, Eligibility, and Results sections. In contrast, SciFive achieves a higher f1-score on hypotheses related to the Adverse Events section. Note that hypotheses concerning the Adverse Events section generally require more numerical calculations than those related to other sections. These results demonstrate that SciFive enhances the system's performance with its numerical inference ability.
\section{Conclusion}
We propose a Multi-granularity system for CTR-based textual entailment and evidence retrieval. The system comprises an ensembling of inference networks and a joint inference module to enhance prediction consistency. A Multi-granularity Inference Network (MGNet) is proposed to handle the two subtasks within a unified network. Another text-generation-based network SciFive is also leveraged to enhance the numerical inference of the system. We use model ensembling to integrate different models' output and enhance the inference stability, while a joint inference method is further proposed to increase the consistency of inference. Our system achieves the best performance on both subtasks. 
The ablation experiments and analysis indicate the effectiveness of different parts of the system. In the future, we will try other large language models and new inference methods (e.g, in-context learning) to inject more medical knowledge into the system and further improve the performance on this task.



\bibliography{anthology,custom}
\bibliographystyle{acl_natbib}
\clearpage
\appendix
\section{Details of Data}\label{sec:data details}
Table \ref{table:basic statistics of NLI4CT} shows the basic statistics of NLI4CT. Avg Hypo denotes the average number of words in hypotheses, Avg CTR Sents refers to the average number of sentences in CTRs, and Avg CTR Words is the average number of words in CTRs. 
\begin{table}[h]
\centering
\setlength{\tabcolsep}{1mm}{
	\begin{tabular}{ccccc}
\hline
\multirow{2}{*}{Split} & \multirow{2}{*}{\#Hypothesis} & \multirow{2}{*}{Avg Hypo} & \multicolumn{2}{c}{Avg CTR}                           \\ \cline{4-5} 
                       &                               &                           & \multicolumn{1}{l}{Sents} & \multicolumn{1}{l}{Words} \\ \hline
Train                  & 1,700                         & 19.7                      & 16.9                      & 165.8                     \\
Val                    & 200                           & 18.7                      & 16.2                      & 166.3                     \\
Test                   & 500                           & 21.6                      & 14.1                      & 161.7                     \\ \hline
\end{tabular}}
\caption{The basic statistics of NLI4CT. }
	\label{table:basic statistics of NLI4CT}
\end{table}

\begin{figure}[h]
	\centering
	\includegraphics[scale=0.35]{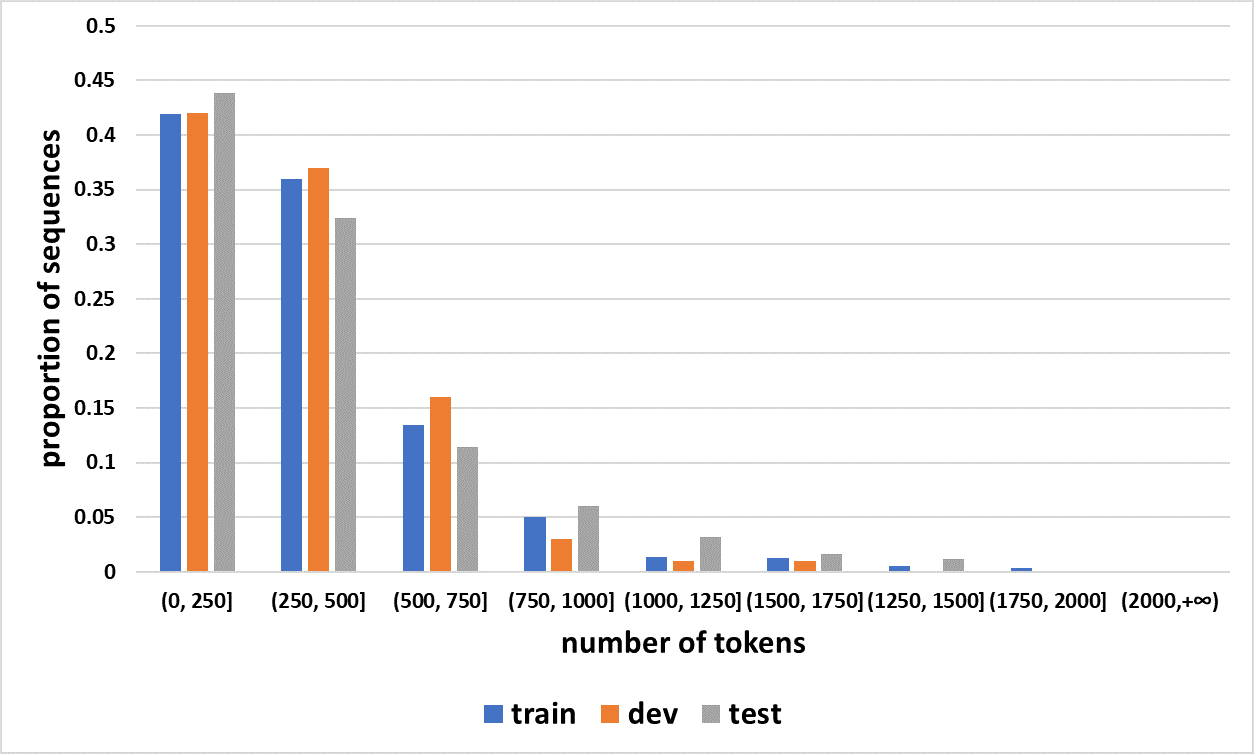}
	\caption{The proportion of input sequences regarding the number of tokens in sequences.}
	\label{fig:statistics}
\end{figure}

Fig.\ref{fig:statistics} depicts the distribution of the input sequences (concatenation of the hypothesis with premise) regarding the number of tokens in sequences. Most input sequences have less than 500 tokens, while there also exists input sequence longer than 1750 tokens.

As for the CTR, each CTR contains a total of four sections:\footnote{The definition of each section is from \url{https://sites.google.com/view/nli4ct/home}.}
\begin{itemize}
    \item Intervention: Information concerning the type, dosage, frequency, and duration of treatments being studied.
    \item Eligibility: A set of conditions for patients to be allowed to take part in the clinical trial
    \item Results: Number of participants in the trial, outcome measures, units, and the results.
    \item Adverse Events: These are signs and symptoms observed in patients during the clinical trial.
\end{itemize}
It is worth to note that each hypothesis only involves a single section in CTR. For hypotheses that involve multiple CTRs, one is required to compare the same section of different CTRs.
\section{Data Preprocessing}\label{sec:input}
We process the evidence CTRs differently for Task A and B. For hypotheses that require single CTR, we simply concatenate all the sentences in the CTR to form the premise sequence for Task A and B. For hypotheses that require comparison between two CTRs, we concatenate the first and second CTR in a single sequence for Task A with the following format: ``primary trial: [CTR 1]. secondary trial: [CTR 2]." Therefore, the joint semantics between two CTRs can be captured by the language model. Nevertheless, we find that encoding each CTR with the hypothesis achieves higher performance for Task B. Therefore, we conduct evidence retrieval on each CTR individually.
\section{Details of Training Joint Inference Network}\label{sec:JIN}
It is hard to train the joint inference network on NLI4CT, because most of the premise corresponds to a pair of hypotheses that contradicts with each other and we can't find positive samples (hypotheses that share the same label). To address this problem, we use the back-translate method to generate positive samples. For each hypothesis $S$ in the train and dev set, we translate it to Chinese and back translate to English to get a new hypothesis $S'$. Then $S$ and $S'$ share the same label for Task A. For a pair of conflicting hypotheses $S_1, S_2$, we generate four input sequences for the joint inference network: 
\begin{enumerate}
    \item $[[\mathrm{CLS}],\textbf{S}_1,[\mathrm{SEP}], \textbf{S}'_1,[\mathrm{SEP}],\textbf{P},[\mathrm{SEP}]]$
    \item $[[\mathrm{CLS}],\textbf{S}_2,[\mathrm{SEP}], \textbf{S}'_2,[\mathrm{SEP}],\textbf{P},[\mathrm{SEP}]]$
    \item $[[\mathrm{CLS}],\textbf{S}_1,[\mathrm{SEP}], \textbf{S}_2,[\mathrm{SEP}],\textbf{P},[\mathrm{SEP}]]$
    \item $[[\mathrm{CLS}],\textbf{S}_2,[\mathrm{SEP}], \textbf{S}_1,[\mathrm{SEP}],\textbf{P},[\mathrm{SEP}]]$
\end{enumerate}
After the data generation, we train the joint inference network on the train set and choose the best model on the dev set. We set the learning rate, warmup ratio, and batch size as 2e-5, 0.1 and 32 respectively.
\end{document}